\documentclass[screen,acmsmall,review=false]{acmart}

\AtBeginDocument{%
  \providecommand\BibTeX{{%
    \normalfont B\kern-0.5em{\scshape i\kern-0.25em b}\kern-0.8em\TeX}}}

\setcopyright{acmcopyright}
\acmJournal{TOMM}
\acmYear{2021} \acmVolume{1} \acmNumber{1} \acmArticle{1} \acmMonth{1} \acmPrice{15.00}\acmDOI{10.1145/3491228}

\acmJournal{TOMM}
\acmPrice{}
\acmISBN{}
\usepackage{bbding}
\usepackage{pifont}
\usepackage{wasysym}

\usepackage{amssymb}
\usepackage{multirow}
\usepackage{soul}
\soulregister\cite7 
\soulregister\citep7 
\soulregister\citet7 
\soulregister\ref7 
\soulregister\pageref7 

\begin{document}

\title{Skeleton Sequence and RGB Frame Based Multi-Modality Feature Fusion Network for Action Recognition}

\author{Xiaoguang Zhu}
\email{zhuxiaoguang178@sjtu.edu.cn}
\orcid{0000-0001-9554-2133}
\affiliation{%
  \institution{Shanghai Jiao Tong University}
  \streetaddress{800 Dongchuan Rd}
  \city{Minhang}
  \state{Shanghai}
  \country{China}
  \postcode{200240}
}

\author{Ye Zhu}
\email{yzhu96@hawk.iit.edu}
\orcid{}
\affiliation{%
  \institution{Illinois Institute of Technology}
  \streetaddress{10 West 31st Street}
  \city{Chicago}
  \state{Illinois}
  \country{U.S.A.}
  \postcode{60616}}

\author{Haoyu Wang}
\orcid{0000-0002-4314-6099}
\affiliation{%
 \institution{Shanghai Jiao Tong University}
 \streetaddress{800 Dongchuan Rd}
 \city{Minhang}
 \state{Shanghai}
 \country{China}
 \postcode{200240}}
\email{gogowhy@sjtu.edu.cn}

\author{Honglin Wen}
\orcid{0000-0002-4314-6099}
\affiliation{%
 \institution{Shanghai Jiao Tong University}
 \streetaddress{800 Dongchuan Rd}
 \city{Minhang}
 \state{Shanghai}
 \country{China}
 \postcode{200240}}
\email{linlin00@sjtu.edu.cn}

\author{Yan Yan}
\orcid{}
\email{yyan34@iit.edu}
\affiliation{%
  \institution{Illinois Institute of Technology}
  \streetaddress{10 West 31st Street}
  \city{Chicago}
  \state{Illinois}
  \country{U.S.A.}}

\author{Peilin Liu}
\orcid{0000-0002-5321-2336}
\affiliation{%
  \institution{Shanghai Jiao Tong University}
  \streetaddress{800 Dongchuan Rd}
  \city{Minhang}
  \state{Shanghai}
  \country{China}
  \postcode{200240}}
\email{liupeilin@sjtu.edu.cn}

\renewcommand{\shortauthors}{X.Zhu et al.}
\begin{abstract}
Action recognition has been a heated topic in computer vision for its wide application in vision systems. Previous approaches achieve improvement by fusing the modalities of the skeleton sequence and RGB video. However, such methods have a dilemma between the accuracy and efficiency for the high complexity of the RGB video network. To solve the problem, we propose a multi-modality feature fusion network to combine the modalities of the skeleton sequence and RGB frame instead of the RGB video, as the key information contained by the combination of skeleton sequence and RGB frame is close to that of the skeleton sequence and RGB video. In this way, the complementary information is retained while the complexity is reduced by a large margin. To better explore the correspondence of the two modalities, a two-stage fusion framework is introduced in the network. In the early fusion stage, we introduce a skeleton attention module that projects the skeleton sequence on the single RGB frame to help the RGB frame focus on the limb movement regions. In the late fusion stage, we propose a cross-attention module to fuse the skeleton feature and the RGB feature by exploiting the correlation. Experiments on two benchmarks NTU RGB+D and SYSU show that the proposed model achieves competitive performance compared with the state-of-the-art methods while reduces the complexity of the network.
\end{abstract}



\maketitle

\section{Introduction}
Action recognition, aiming to identify the action category of a person or group in one temporal clip, plays a vital role in industrial applications on Internet of Things and human-robot interaction. 
Recent years have witnessed the increasing popularity in deep-learning based action recognition, of which the data source from two modalities achieves the highest performance: RGB video modality~\cite{longterm_rnn,I3D,dRNN,ntu_rgbd,tomm_action_attention} and skeleton sequence modality~\cite{ntu_rgbd,hierarchical_rnn,review_3d_action,C3D}.

\begin{figure*}[h]
    \centering
    \includegraphics[width =\textwidth]{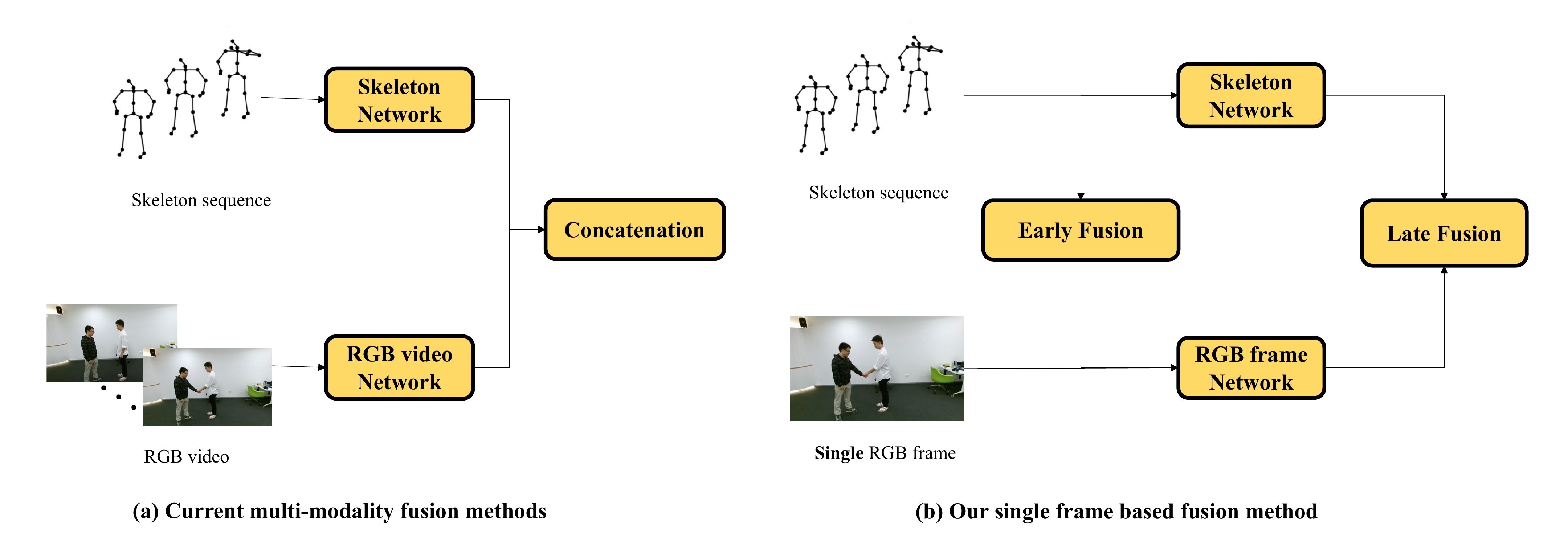}
    \caption{(a) Current multi-modality methods take skeleton sequences and whole RGB videos as input, which conduct simple concatenation or weighted sum in the fusion stage. (b) Our method takes a single RGB frame picked out from the video and performs early and late fusion. The replacement of RGB videos by a single frame maintain the key information as well as save computational workload significantly.}
    \label{fig_first_pic}
\end{figure*}

The skeleton sequence and RGB video have different characteristics in action recognition, as shown in Table~\ref{tab_first_pic}. The skeleton sequence modality contains the movement of the human limbs which is temporal information. And the RGB video modality contains not only temporal information but also abundant spatial information such as the description of the human limbs and the human-object interaction~\cite{tomm_action2,tomm_action_unsupervised}. Thus various approaches are proposed to exploit the information in both modalities. The works that process RGB video modality use Long Short-Term Memory (LSTM) or 3D Convolutional Neural Network (CNN) to explore spatial and temporal information. However, though RGB video also contains temporal information, yet it is less direct or comprehensive compared with the skeleton modality, as it is susceptible to the background and thus the temporal information in RGB video modality is referred to as 'Weak' in Table~\ref{tab_first_pic}. As to the skeleton sequence modality, recent works employ LSTM~\cite{ST-LSTM,TS-LSTM,AGC-LSTM} or Graph Convolutional Network (GCN)~\cite{ST-GCN,As-GCN,CA-GCN} based networks to obtain the temporal movement embeddings, which gain more satisfying and efficient performance than RGB video-based method. However, since the skeleton sequences modality naturally lacks spatial information, it can be difficult to predict the action precisely with human-object interaction. For example, the skeleton sequence with arms up can either wear a cap or drink a bottle of water. Following researchers explore the fusion network to take advantage of the complementarity of skeleton sequence and RGB video~\cite{SGMNet,2streamfusion,bilinear,tomm_action1,tomm_action3}, which have achieved prominent improvement. However, such methods consume much computational workload, as the RGB video is processed with 3D CNN or hierarchical 2D CNN, which currently requires more than 100 GFLOPs for a single clip~\cite{MFN}. In other words, there is a dilemma between the accuracy and efficiency in the cross-modality methods.

\begin{table*}
\caption{The table illustrates the information contained in different modalities. $\checkmark$ indicates that the information is contained in the modality, while × represents that the information is not. Weak or Strong indicates whether it is easy to obtain the information in the modality.}
\begin{tabular}{ccc}
\toprule
Modality & Temporal Information & Spatial Information  \\ \midrule
RGB Video & Weak & \Checkmark \\ 
Skeleton Sequence & Strong & \XSolidBrush \\ 
RGB Image & \XSolidBrush & \Checkmark \\ \midrule
RGB Video + Skeleton Sequence &  Strong & \Checkmark \\ 
RGB Image + Skeleton Sequence & Strong & \Checkmark \\ \bottomrule
\end{tabular}
  \label{tab_first_pic}
\end{table*}

To address this issue, we propose the Multi-Modality Feature Fusion (MMFF) network to fuse the skeleton sequence and RGB frame modalities for action recognition. Take a deep thought into the information contained in various modalities, the combination of skeleton sequences and a single RGB frame can cover strong temporal information and spatial information, which is close to the combination of the whole RGB video with the skeleton sequence to some extent. Also, considering the situation that the object (bottles, caps, etc.) which interacts with the human mostly occurs in most videos from the beginning to the end, we extract a single middle RGB frame from the video to obtain the spatial information. Notably, the network that extracts the feature of the RGB frame only needs a basic 2D CNN, of which the network complexity is much lower than that of the RGB video network. In this way, the network maintains most key information as well as significantly reducing the network complexity.

However, the replacement of the RGB video by a single frame loses the temporal information in the RGB stream, which makes the skeleton stream and the RGB stream hardly share coincident features. Therefore, we introduce a novel two-stage feature fusion method to better fuse the two modalities. Apart from the previous multi-modality fusion methods which use concatenation or weighted sum~\cite{2streamfusion,chained} as illustrated
in Fig.~\ref{fig_first_pic}(a), the proposed method use attention mechanisms to enhance the correspondence of the two modalities in early and late stages as shown in Fig.~\ref{fig_first_pic}(b). At the early fusion stage, we conduct projection from the skeleton sequence to the RGB frame as an attention mask to guide the RGB network to focus on the most informative regions involving the movement of limbs. Besides, self-attention is also introduced in the RGB frame network to suppress the information of the background. As to the late fusion stage, we perform a cross-attention fusion module to combine the skeleton features with the RGB features together.

We conduct experiments on two popular benchmarks: NTU RGB+D~\cite{ntu_rgbd} and SYSU~\cite{sysu} datasets to verify the effectiveness of the proposed method. Compared with the other state-of-the-art multi-modality methods, our method exceeds them in both performance and computational consumption. In detail, the proposed method outperforms SGM-Net~\cite{SGMNet} by $0.7\%$ under Cross-Subject evaluation on NTU RGB+D with only $40.6\%$ of its parameters size and $14.5\%$ of its FLOPs.

In conclusion, our main contributions are as follows:

$\bullet$ To combine the multi-modality information, we fuse a single RGB frame containing the human-object interactions instead of disposing of the whole RGB video with the skeleton sequence, which significantly reduces the computational consumption as well as maintains the performance.

$\bullet$ We perform a novel two-stage feature fusion network to combine the knowledge of the RGB and the skeleton modalities, in which two attention mechanism components are introduced to help the network concentrate on the human-object interaction regions and explore the correspondence.

$\bullet$ Comprehensive experiments are conducted on NTU RGB+D and SYSU datasets. The ablation study demonstrates the effectiveness of the proposed method. The results show that the network achieves competitive performance compared with other state-of-the-art methods and reduces the network complexity.

The rest of this article is organized as follows, Section 2 explores the related work. Section 3 introduces the proposed multi-modality fusion network. Section 4 describes our implementation details and experiment results. Section 5 provides the conclusion of our article.

\section{Related Work}
\textbf{3D Skeleton Action Recognition.} 3D skeleton-based action recognition has attracted increasing attention in recent years. Due to its high-level representation with background-robust and view-invariant features, the 3D skeleton based method has boosted the performance of action recognition in recent years. According to the type of neural networks, current skeleton-based methods can be mainly divided into three categories: Recurrent Neural Network (RNN), Convolutional Neural Network (CNN) and Graph Convolutional Network (GCN) based methods.

The RNN-based approaches directly take the 3D skeleton joints coordinates as sequence input and then use RNN to memorize the movement of actions~\cite{ntu_rgbd,hierarchical_rnn,hierarchical,memory,independentRNN}. \cite{hierarchical} uses separated RNN sub-nets to extract skeleton features from five human parts. TS-LSTM~\cite{TS-LSTM}, ST-LSTM~\cite{ST-LSTM} and dRNN~\cite{dRNN} modify the Long Short-Term Memory (LSTM) to explore spatial-temporal information. Wang~\textit{et al.}~\cite{temporal_spatial} propose a novel two-stream RNN architecture to model both temporal dynamics and spatial configurations for skeleton based action recognition. Li~\textit{et al.} ~\cite{independentRNN} propose a new type of RNN to solve the problem of gradient exploding and vanishing and make it possible and more robust to build a longer and deeper RNN for high semantic feature learning.

CNN has shown great success in computer vision tasks for its powerful capability of extracting image features. Since the skeleton movement of human action is a temporal sequence, researchers encode the skeleton joints into multiple 2D pseudo-images, and then feed them into CNN to learn useful features~\cite{Ke_2017_CVPR,shape-motion,treeCNN,cuboid}. Ke~\textit{et al.}~\cite{Ke_2017_CVPR} propose to use deep CNN to learn long-term temporal information of the skeleton sequence from the frames of the generated clips. Li~\textit{et al.}~\cite{shape-motion} apply a multi-stream CNN model to extract and fuse deep features from the designed complementary shape-motion representations. Zhu~\textit{et al.}~\cite{cuboid} organize the pairwise displacements between all body joints to obtain a cuboid action representation and use attention-based deep CNN models to focus analysis on actions. 

Inspired by the fact that the skeleton data is naturally a topological graph, where the joints and bones are regarded as the nodes and edges, Graph Convolutional Network (GCN) is adopted to boost the performance of skeleton based action recognition~\cite{ST-GCN,As-GCN,DGNN,2s-AGCN,MS-G3D,CA-GCN}. ST-GCN~\cite{ST-GCN} firstly constructs the skeleton as a graph and uses GCN to automatically capture the patterns embedded in the spatial configuration of the joints as well as their temporal dynamics. As-GCN~\cite{As-GCN} extends the skeleton graph by integrating the structural links and action links and learn both spatial and temporal features for action recognition. CA-GCN~\cite{CA-GCN} considers a context term for each vertex by integrating information of all other vertices and simplifies the network greatly. DGNN \cite{DGNN} and MS-G3D \cite{MS-G3D} develop various forms to construct the body graph and fuse the joint-bone features for action recognition.

\textbf{Multi-Modality Action Recognition.} Since different modalities contain different complementary information, the fusion of multiple feature streams has been widely exploited and proved to boost classification performance. 
Earlier works focus on late fusion strategy using score fusion~\cite{2streamfusion,Zhang_TMM,Wu_ACMMM,mtda,joule}, while most current methods tend to design correlation module for modality fusion \cite{posemap,bilinear,SGMNet,Shahroudy_TPAMI,MMTM,mfas,2d3d_pose,JOLO-GCN}. Zhao~\textit{et al.}~\cite{2streamfusion} propose a two-stream network to process the skeleton and video feature separately and use the SVM scoring method for fusion. Liu~\textit{et al.}~\cite{posemap} use a two-stream network to extract features from the skeleton and pose heatmaps, and concatenate the two features for fusion and final prediction. Li~\textit{et al.}~\cite{SGMNet} develop the bilinear pooling~\cite{bilinear} to explore the multi-modality correlations. Joze~\textit{et al.}~\cite{MMTM} propose to fuse the multi-modality features in different spatial dimensions, which utilizes the knowledge of multiple modalities to re-calibrate the channel-wise features in each CNN stream. Cai~\textit{et al.} \cite{JOLO-GCN} propose to employ human pose skeleton and joint centered light-weight information jointly in a two-stream
graph convolutional network.

\textbf{Attention Mechanism.} The attention mechanism is firstly used in image caption~\cite{show_attend_tell} and then applied in many other fields. In the action recognition task, the attention mechanism has also been exploited and achieves large improvement~\cite{GCA-LSTM,ST-attention,AGC-LSTM,ANT,JCAGCN,SDAN}. Liu~\textit{et al.}~\cite{GCA-LSTM} propose the GCA-LSTM to selectively focus on the informative joints in the action sequence with the assistance of global contextual attention. Song~\textit{et al.}~\cite{ST-attention} propose a spatial and temporal attention model to explore the discriminative features of key joints and frames for human action recognition from skeleton data. The attention mechanism is employed in AGC-LSTM~\cite{AGC-LSTM} to enhance the information of key joints. Ji~\textit{et al.}~\cite{ANT} propose to transfer attention from the reference view to arbitrary views, which correctly emphasizes crucial body joints and their relations for view-invariant representation. Chen~\textit{et al.}~\cite{JCAGCN} incorporate the joint-wise channel attention with the GCN to mine discriminative information among confusing actions.

Taking a deep insight into these works, we propose to combine the skeleton modality with only one RGB frame from the action video modality for action recognition. Compared with other multi-modality models, the proposed method can reduce complexity by a large margin. Specifically, a self-attention module is proposed to focus on the human body and the human-object interaction region, and a skeleton attention module is proposed to project the skeleton onto the RGB frame to guide the network focus more on the human limb movement regions. Moreover, a multi-modality fusion module is used to fuse the feature of skeleton and image modalities by exploring the correlations, which differs from previous methods that use feature concatenation and pose-guided information.

\section{Methodology}
The key idea of the proposed MMFF is to combine the temporal feature in the skeleton modality with the spatial feature in the single RGB frame. In this section, we first present an overview of the method. We then briefly review the ST-GCN~\cite{ST-GCN} and Bi-LSTM~\cite{BiLSTM} networks, which are the backbones of the skeleton stream. Next, we present the data enhancement technique. Finally, the RGB stream and the two-stage multi-modality fusion module are introduced. 

\subsection{Overview}
\begin{figure*}[h]
    \centering
    \includegraphics[width =4.5in]{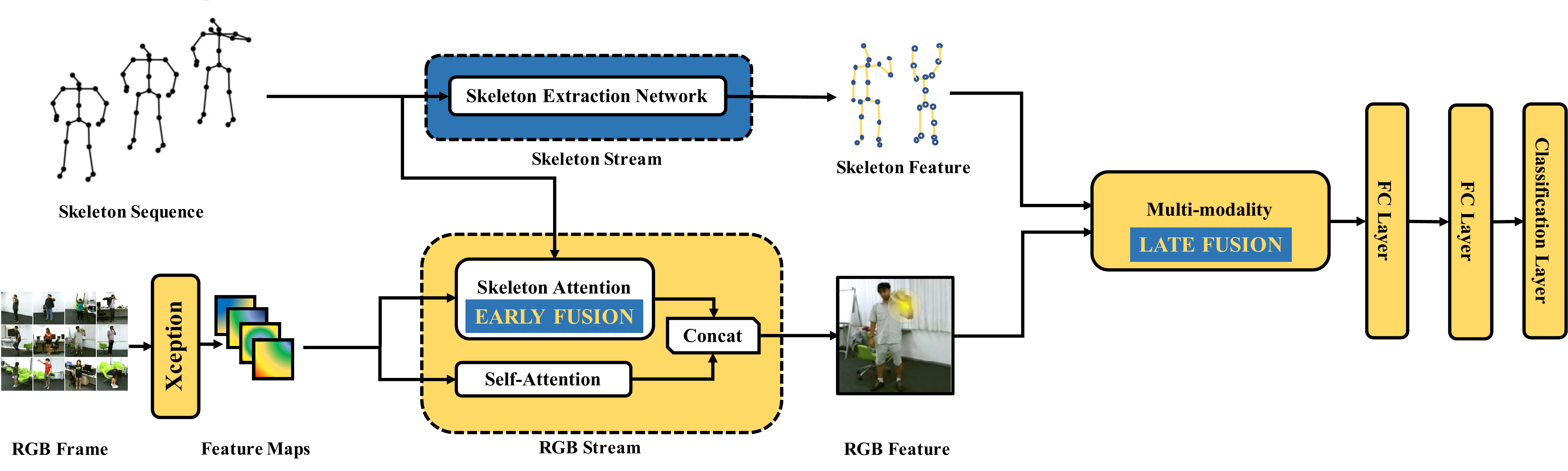}
    \caption{The overall architecture of our model has two inputs sources: one skeleton sequence source and one single RGB frame picked from the video. In the skeleton stream, LSTM/ST-GCN is used to process the skeleton sequence modality and generate the skeleton feature map}. In the RGB stream, the skeleton is projected to serve as the guide of the limb movement position attention module as well as the first feature fusion stage. A self-attention module is also implemented to focus on the foreground regions. Both the outputs of the attention module are concatenated together to generate the RGB feature map. The skeleton feature map and the RGB feature map are fused through the late fusion module to obtain the final result.
    \label{overall_architecture}
\end{figure*}
The skeleton and RGB modalities focus on different key information for action recognition. As the skeleton modality is a sequence with joints and bone movement features, the skeleton feature contains temporal features such as the movement of human limbs. Due to the lack of RGB pixels, the skeleton modality lacks spatial information such as the description of the limb and human-object interactions. 
As to the RGB modality, its advantages lie in the abundant spatial information mentioned above. Though the RGB modality also contains temporal information, it is far less direct or comprehensive compared with the skeleton sequences. Hence it is beneficial to combine these two complementary modalities for action recognition. The middle frame of the RGB video is picked as the RGB frame in the network, yet in Section 4 we verify that the frames from the middle part of the video are also effective.

As illustrated in Fig.~\ref{overall_architecture}, the architecture of the proposed method contains three parts: the Skeleton Stream, the RGB Stream and the Feature Fusion Module. In the skeleton stream, the skeleton sequence is fed into the feature extraction network. In this article, we use the ST-GCN~\cite{ST-GCN} or triple-layer Bi-LSTM~\cite{BiLSTM} as the backbone to extract the skeleton feature for a fair comparison with the SOTA methods in~\cite{SGMNet} and~\cite{2streamfusion}. For the RGB frame, the Xception~\cite{Xception} network is used to extract the RGB feature. Specifically, in RGB stream, two attention components are performed: the self-attention module is used to help the network focus on the human body parts, while the skeleton attention projects the skeleton feature onto the RGB frame, serving as the early feature fusion stage to help the network focus on the human limb movement regions in the RGB frame. In the final stage, the skeleton feature and the RGB feature are fused through the late-stage multi-modality fusion module.

\subsection{The Skeleton Stream}
In the skeleton stream, we use two different models as the backbone network to extract the skeleton sequence feature. The ST-GCN~\cite{ST-GCN} is the GCN-based skeleton method, while the stacked Bi-LSTM~\cite{BiLSTM} is the RNN-based method.

\textbf{ST-GCN.} In ST-GCN, the skeleton sequence is constructed as an undirected spatial temporal graph $G=(V,E)$ with N joints and T frames featuring intra-body and inter-frame connection. In the graph, the nodes contain all the skeleton joints and are represented as $V= \left\{v_{ti}|t=1,2,...,T,i=1,2,...,N \right\}$. The edge set is composed of two subsets, of which one is the connectivity of human body structure within one frame, and the other one is the connectivity of each joint to the same joint in the consecutive frame. The ST-GCN adopts a similar graph convolution operation in~\cite{GCN} to extract the spatial temporal features:
\begin{equation*}
    f_{out}=\sum_{j}{\Lambda_j^{-\frac{1}{2}}A_j\Lambda_j^{-\frac{1}{2}}f_{in}W_j},
\end{equation*}where ${A}_j$ is the dismantled matrix of original adjacency matrix A with ${A}+{I}=\sum\nolimits_{j}{{A}_j}$ and ${I}$ is the identity matrix. $\Lambda_j^{-\frac{1}{2}}A_j\Lambda_j^{-\frac{1}{2}}$ is the normalization of adjacency matrix $A_j$ using degree matrix $\Lambda_j$, where $\Lambda_j^{ii}=\sum\nolimits_{k}(A^{ik}+I^{ik})$. $W_j$ is a learnable weight matrix, $f_{in}$ is the input graph feature and $f_{out}$ is the output feature map with the shape of $C_s\times N \times T$. 

\textbf{Bi-LSTM.} LSTM~\cite{LSTM} can handle sequential data within various steps. Compared with RNN, LSTM can learn long-range dependency~\cite{Vemulapalli_2014_CVPR} and avoid the problem of gradient vanishing. A typical LSTM neuron contains an input gate $i_t$, a forget gate $f_t$, 
a cell state $c_t$, an output gate $o_t$, and an output response $h_t$. The LSTM transition equations can be expressed as:
\begin{equation}
\begin{pmatrix}
i_t\\
f_t\\
o_t\\
u_t
\end{pmatrix}=
\begin{pmatrix}
sigm\\
sigm\\
sigm\\
tanh
\end{pmatrix}
W
\begin{pmatrix}
x_t\\
h_t-1\\
\end{pmatrix}
,
\end{equation}
\begin{equation}
    c_t = f_t \otimes c_{t-1} + i_t \cdot u_t,
\end{equation}
\begin{equation}
    h_t = o_t \cdot tanh(c_t),
\end{equation}{}where $sigm$ and $tanh$ are activation functions of sigmoid and tanh.

In this article, we adopt the Bi-LSTM module with three layers as the backbone of the RNN-based method. The input skeleton stream is denoted by $I \in R^{N\times 2\times T}$, where N is the number of the skeleton points in the $t$th frame, T is the number of the total time steps in the skeleton sequence. For each time step, the input information is $x_t \in \mathbb{R}^{N \times3} $. The output feature is the cell state of the last time step with the shape of $C_s$, which is two times the number of the hidden units in Bi-LSTM.

\subsection{Data Enhancement}

On processing the skeleton sequence and the single RGB frame, the skeleton sequence has only one viewpoint. In addition, the human body in the RGB video only occupies a small part of its whole image. Both situations would restrict the recognition performance, thus we introduce the measures to enhance the skeleton data with data augmentation and the RGB data with projection crop.

\begin{figure}[t]
    \centering
    \includegraphics[width =0.95\textwidth]{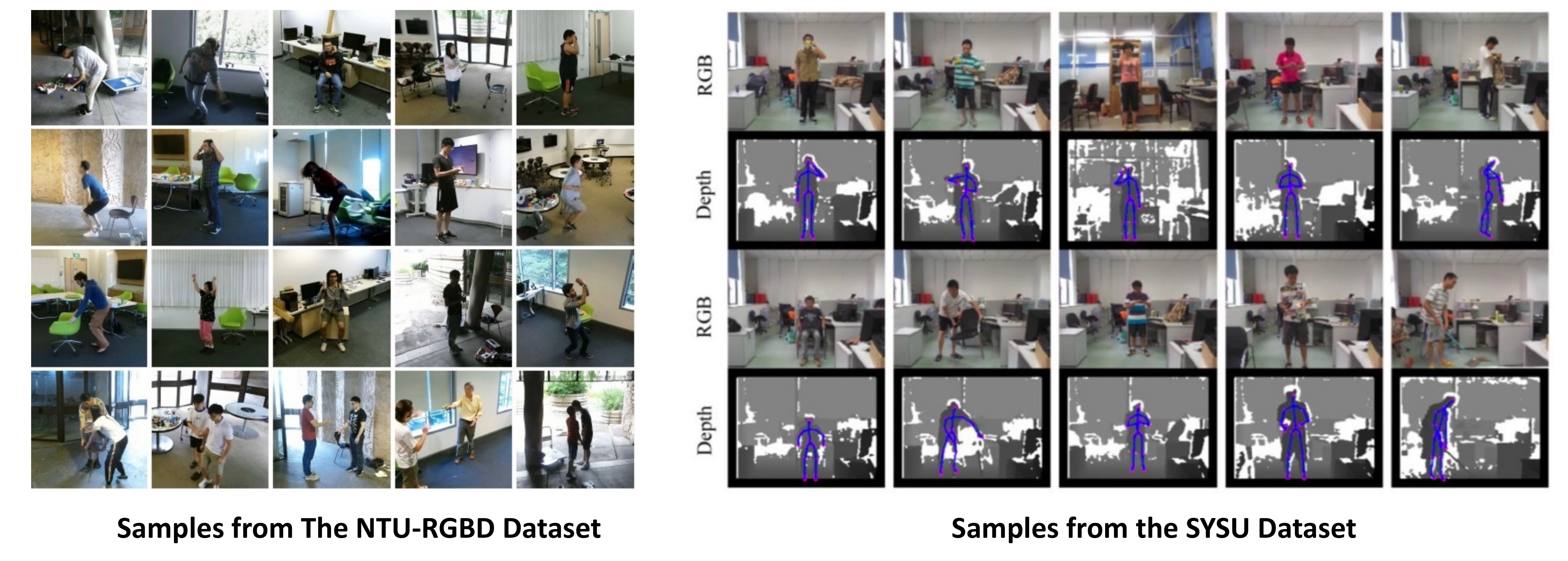}
    \caption{The samples selected from the NTU RGB+D and the SYSU dataset. The NTU RGB+D provides the individuals from multiple views like $0^{\circ}$ +$45^{\circ}$ and $-45^{\circ}$ against the camera. The SYSU dataset contains the individuals $0^{\circ}$ right against the camera.}
    \label{fig_6}
\end{figure}

\textbf{Data Augmentation.} The data in our implementation is combined with data augmentation based on the rotation matrix and scale-changing. As shown in Fig.~\ref{fig_6}, the NTU RGB+D and SYSU datasets have limited viewpoints. In the NTU RGB+D dataset, the viewpoint of the skeleton various in 3 different degrees: $0^{\circ}$ against the camera, +$45^{\circ}$ and minus $45^{\circ}$ against the camera. In the SYSU dataset, there only exists the data which is 0 degree against the camera. To increase the robustness, we enhance the data by rotating the matrix. The rotation matrix formula is as follows, where $\alpha$, $\beta$ and $\gamma$ stand for the rotation degree in the clockwise direction in x, y, z axis.
\begin{equation}
R_x(\alpha) =  
\left[
 \begin{matrix}
   1 & 0& 0\\
   0&\cos{\alpha}& -\sin{\alpha} \\
   0 & \sin{\alpha} & \cos{\alpha}
  \end{matrix}
 \right],
\end{equation}
\begin{equation}
R_y(\beta) =  
\left[
 \begin{matrix}
   \cos{\beta} & 0& \sin{\beta}\\
   0&1& 0 \\
  -\sin{\beta} & 0& \cos{\beta}
  \end{matrix}
 \right],
\end{equation}

\begin{equation}
R_z(\gamma) =  
\left[
 \begin{matrix}
   \cos{\gamma} & -\sin{\gamma}& 0\\
   \sin{\gamma}&\cos{\gamma}& 0 \\
  0 & 0& 1
  \end{matrix}
 \right].
\end{equation}
After multiplying the three rotation matrices, we get:
\begin{equation}
    \centering
    R = R_z(\gamma)R_y(\beta)R_x(\alpha).
\end{equation}{}In our implementation, $\alpha$ and $\beta$ vary between $0^{\circ}$  and $30^{\circ}$. Moreover, we involve the 'Change of scale' method to further increase the variety:
\begin{equation}
    \centering
    S = 
    \left[
 \begin{matrix}
   S_x & 0& 0\\
  0&S_y& 0 \\
  0 & 0& S_z
  \end{matrix}
 \right],
\end{equation}{}where $s_x$, $s_y$ and $s_z$ are the scale-changing factors along x,y and z axis. In our implementation, $s_x$, $s_y$ vary between 1 and 1.2, extending the datasets to five times bigger. Through the matrix rotation method and the scale-changing method, the variety of the original dataset gets boosted.

\textbf{Projection Crop.} In our chosen RGB frame, human only occupies a small region of the whole image. To tackle this problem, we propose a projection crop method to preprocess images. This procedure crops the part of the human subject according to the bounding box, which is according to the corresponding skeleton data of this frame. Assume that $w$ and $h$ are the width and height of the bounding box, we use four corners of the bounding box as the origin and crop the original image with $w + w'$ and $h + h'$ where $w'$ and $h'$ ranging from 100 to 300 pixels randomly. Thus the number of images is augmented by 4 times. 

Through the projection crop method, given the 3D skeleton coordinates and the camera parameters, the 3D skeleton sequence is projected onto the RGB frame. Then the relationship between the skeleton sequence and the RGB frame can be detected and the 2D pixel coordinates can be calculated through the projection equation. Finally, the region of the human subject can be determined. Overall, the proposed method is a variant to the commonly used crop methods in image or video based tasks. The proposed method crops the images with the human as center prior, while the other methods use the image center for random crop. As the proposed crop method is compatible with the skeleton attention to make the coordinates of the two modalities the same, it can be regarded as part of the whole method itself. The advantages of the proposed projection are twofold: (1) The human parts take up most of the picture, which mitigates the impact of the background; (2) The center of the images is fixed on the human, which makes the skeleton attention easy to be applied to the image. However, as the other fusion-based action recognition methods use RGB video for spatial information extraction, it would be far more complex to perform projection crop to each frame and the alignment of coordinates for the frames may affect the integrity of the video representation.

\subsection{The RGB Stream}
In our implementation, the RGB stream consists of three parts: the base convolution layers, the self-attention module and the skeleton attention module. The Xception~\cite{Xception} network is used as the base convolution layers to extract the feature maps. The self-attention module and the skeleton attention module are used to generate attention weights.

\begin{figure}[t]
    \centering
    \includegraphics[width =3.5in]{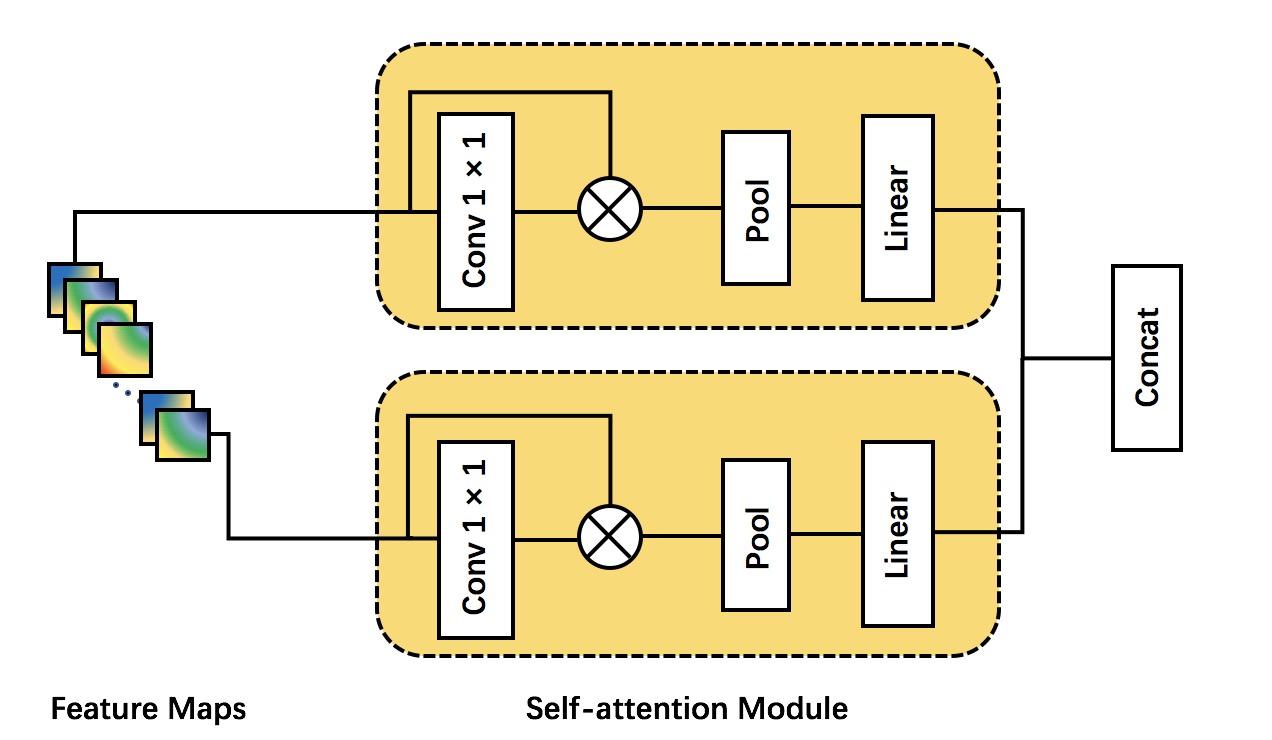}
    \caption{Self-attention module. The module takes the feature map extracted from the RGB frame by Xception network. The self-attention module has 2 repeated branches of the dashed line boxes, which are used to generate the weights of the attention module. In the dashed line box, '$1 \times 1$ Conv' denotes the $1 \times 1$ convolutional layer, $\otimes$ means the matrix multiplication, 'Pool' layer stands for global average pooling, and 'Linear' layer stands for liner transformation which reduces the dimension to 256. The output of each self-attention box has a dimension of 256, so after the 'Concat' layer, the output dimension reaches to 512.}
    \label{fig_2}
\end{figure}
\textbf{Self-Attention Module.} Inspired by the method of extracting the body part features in person re-identification~\cite{Zhao_2017_ICCV}, we proposed a self-attention module to conduct the feature maps of the original RGB frame. The self-attention module aims to extract the visual saliency of an action from the RGB frame, i.e., to emphasize the features of the human part and suppress the features of the background. The self-attention module is different from the original soft attention module used in the image caption assignments~\cite{show_attend_tell} as follows:

\begin{equation}
e_{ti} = f_{att} (a_i,h_{t-1}),
\end{equation}
\begin{equation}  
 \alpha_{ti} = \frac{\exp{e_{ti}}}{\sum_{i=1}^L\exp{e_{ti}}},
\end{equation}
where $f_{att}$ denotes the attention module, which is a multilayer perceptron. $t$ represents the time steps while $i$ represents the distinct image locations. $a_i$ denotes the image feature and $h_{t-1}$ is the hidden context state of the previous time step. $a_{ti}$ is the final attention weight, which is a probability between 0 and 1 obtained with $e_{ti}$ going through the softmax layer.

In the action recognition task, we do not have context information as in the image caption assignments. Therefore, the hidden states $h_{t-1}$ and the time step $t$ are not available. As shown in Fig.~\ref{fig_2}, an $1 \times 1$ convolution layer $Conv$ is used to replace the $f_{att}$ attention module. For the input feature map $F_{in}\in\mathbb{R}^{C \times{W} \times{H}}$, where $W$ and $H$ denote the width and height of feature map, the $1 \times 1$ convolution is used to generate the attention mask $M_{self}\in \mathbb{R}^{1 \times{W} \times{H}}$. A~\emph{sigmoid} function $\sigma$ is used to replace the softmax function to concentrate the attention weight probability in the range of $[0,1]$:
\begin{equation}
M_{self} = \sigma(Conv(F_{in})).
\end{equation} Then the output feature $F_{self}\in \mathbb{R}^{C \times{W} \times{H}}$ is calculated as:
\begin{equation}
F_{self}^{RGB} = M_{self}\odot F_{in},
\end{equation}where $\odot$ denotes the element-wise multiplication.

\textbf{Skeleton Attention Module.} In this module, a single RGB frame and the skeleton sequence are combined as the early fusion stage. As the static RGB frame does not contain temporal information, it is sufficient to use the skeleton sequence to guide the image to focus on the moving human-object region as a supplement.
\begin{figure}[t]
    \centering
    \includegraphics[width =0.9\textwidth]{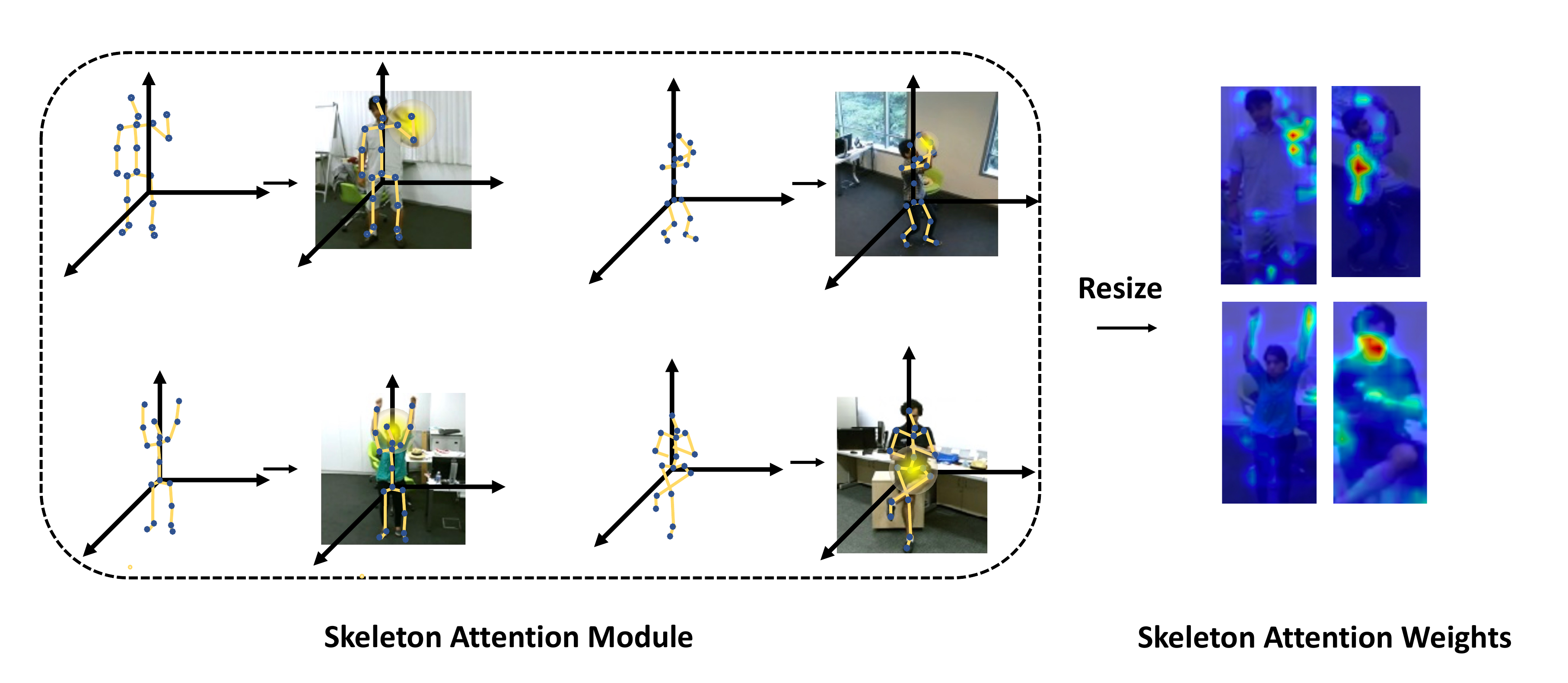}
    \caption{The procedure of the generation of the skeleton attention weights. The first coordinate system stands for the skeleton sequence and the second one stands for the middlemost RGB frame. The skeleton sequence is projected to the middlemost RGB frame. The attention module speculates the most interesting parts through this procedure.
    For example, in the particular left first image shown above, through calculating in the skeleton attention module, the left hand of the person has the biggest moving distance. The yellow part of his left hand is the attention mask we added. Then, the final skeleton attention weights are obtained by resizing the attention mask to the cropped image.}
    \label{fig_skeleton_attention}
\end{figure}

At first, the skeleton joint with the largest moving distance $d_{max}$ is computed as:
\begin{equation}
d_{max} = \parallel J_{1,j_{max}} - J_{mid,j_{max}} \parallel_2,
\end{equation}
\begin{equation}  
j_{max} = arg \mathop{\max}_{j} \parallel J_{1,j} - J_{mid,j}\parallel_2.
\end{equation}where $J_1$ and $J_{middle}$ stand for the 3D joint locations of the skeleton frame and the middle RGB frame, respectively. By calculating the moving distance, the index of the joints with the biggest changes $j_{max}$ is picked out. Secondly, we start to generate the skeleton attention mask $M_{ske}\in \mathbb{R}^{1 \times{W} \times{H}}$. When generating the skeleton attention mask, the attention weight $M_{ske}^{p} = 1$ in a square centered at $j_{\max}$ while $M_{ske}^{p} = 0$ at the other locations, $p$ means the pixel location of the mask. Then, the attention mask has the same spatial size as the feature maps is resized, and the skeleton attention weights are generated. At last, the human-object related features $F_{ske}\in \mathbb{R}^{C \times{W} \times{H}}$ is obtained with the element-wise multiplication of the skeleton attention weight and the input feature map $F_{in}$.
\begin{equation}
F_{ske}^{RGB} = M_{ske}\odot F_{in},
\end{equation}where $\odot$ denotes the element-wise multiplication.

The process of generating the skeleton attention can be regarded as a projection from the skeleton sequence to the single RGB frame in a relatively prior stage. As illustrated in Fig.~\ref{fig_skeleton_attention}, the left arm of the person can be emphasized through the skeleton attention mask. In this way, the RGB stream can extract the human-object interactive features with the guide of skeleton movement information.  

\subsection{The Late Fusion Module}
After utilizing the skeleton information to guide the RGB attention, the features of the two sub-network streams (the skeleton stream and the RGB stream) are combined for action classification. Since the traditional decision fusion has too much dependency on the datasets, and inspired by the multiple streams fusion methods ~\cite{Rahmani_2017_ICCV}~\cite{2streamfusion}~\cite{longterm_rnn}, feature fusion is applied to help leverage the complementary information between the two modalities.

As the dimensions of the feature map from the ST-GCN and stacked Bi-LSTM are different, denoted as $F_{LSTM}\in \mathbb{R}^{C_{S}}$ and $F_{GCN}\in \mathbb{R}^{C_{S} \times{T} \times{V}}$ respectively, where $C_{S}$ is the channel dimension, $T$ is the length of time and $V$ is the number of skeleton nodes, we implement LSTM-based fusion and GCN-based fusion in different ways.

\textbf{LSTM-Based Fusion Module.}
As the feature of the skeleton stream is compressed with only dimension $C_S$, the late fusion strategy is applied to fuse the RGB features $F_{RGB}$. The RGB feature is generated by combining the self-attention feature $F_{self}^{RGB}$ and the skeleton attention feature $F_{ske}^{RGB}$. The two features first pass through the MaxPooling layer to obtain features with the dimension of $C$, and then are concatenated to form the feature $F_{RGB}\in \mathbb{R}^{C_R}$ of RGB stream.

To implement the final feature fusion of the skeleton feature and the RGB feature, firstly, we concatenate the features from the two streams. Secondly, we add an $L_2$ normalization layer to the concatenated feature. Then, a fully connected layer with leaky ReLU as the activation function is used. This layer is intended to find the non-linear relationship between temporal skeleton features and spatial RGB features. The last two layers are fully connected softmax layers which are used for classification.  

\begin{figure}[t]
    \centering
 
    \includegraphics[width =5in]{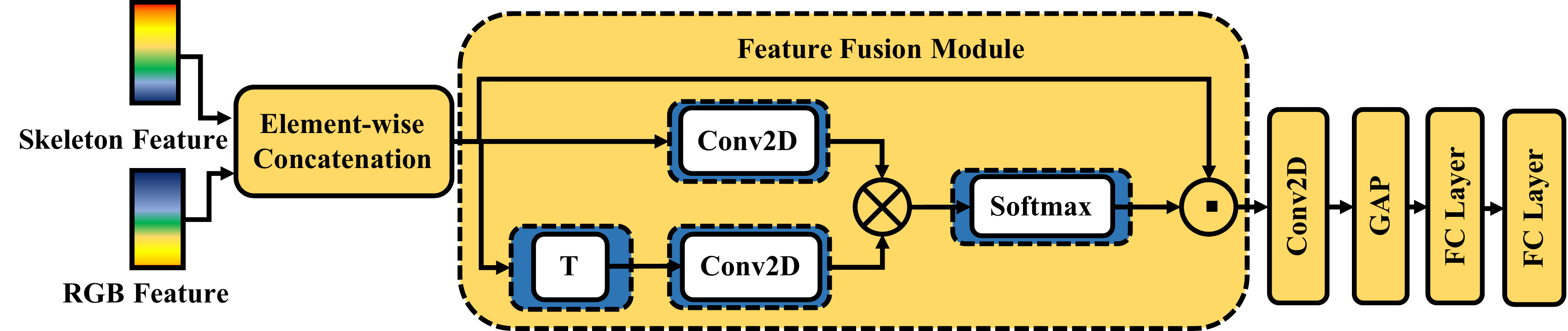}
    \caption{The architecture of our proposed Multi-Modality Feature Fusion (MMFF) Module. The skeleton feature and the RGB feature are element-wise concatenated and then sent into the module. $T$ denotes transpose, $Conv2D$ means 2D convolutional layer, $\otimes$ stands for matrix production and $Softmax$ represents the activation function. The output of the fusion module is sent into one 2D convolutional layer, on global average pooling layer and two FC layers to obtain the final feature embeddings.}
    \label{featurefusion}
\end{figure}

\textbf{GCN-Based Fusion Module.}
To better explore the relation between the two modalities and combine the complementary information, we design a fusion module with full consideration of feature characters. For the skeleton feature $F_{GCN}\in \mathbb{R}^{C_{S} \times{T} \times{V}}$, it contains the temporal-spatial feature, while for the RGB feature $F_{RGB}\in \mathbb{R}^{C_{R} \times{H} \times{W}}$, it only contains the spatial feature. In the fusion module, as the skeleton attention has already helped the RGB stream extract the key information, we focus on exploit the spatial relation across the modalities as illustrated in Fig.~\ref{featurefusion}. 

For the RGB feature, the two attention features are average summed to obtain the RGB feature $F_{RGB} \in \mathbb{R}^{C_R \times S}$, where $S$ is the multiplication of $H \times W$. For the skeleton feature, we use max pooling to transform the feature which contains only spatial feature $F_{GCN} \in \mathbb{R}^{C_S \times V}$. The RGB feature and the skeleton feature are then transformed to vectors $f_{RGB} \in \mathbb{R}^{C_R}$ and $f_{GCN} \in \mathbb{R}^{C_S}$ with Global Average Pooling (GAP). The element-wise concatenation is used to combine the two features. For the RGB feature, the skeleton vector $f_{GCN}$ is concatenated to each channel feature of $F_{RGB}$, while for the skeleton feature, the RGB vector $f_{RGB}$ is concatenated to each channel feature of $F_{GCN}$. Thus the channel dimension of both features are equal, and then the combined feature $F_{com}$ is obtained by concatenating along the channel dimension. The combined feature is in the shape of $(C_S + C_R) \times (S+V)$.

As the combined feature contains the spatial features from both the skeleton stream and the RGB stream, we use two $1 \times 1$ convolutional layers $Conv$ to explore the cross-spatial relation and generate a relation mask $M_{rel}$:
\begin{equation}
M_{rel} = \sigma(Conv(F_{com}) \times Conv(F_{com}^{T})),
\end{equation}where $\sigma$ denotes the softmax function. Then the output feature is obtained as follows:
\begin{equation}
F_{rel} = M_{rel}\odot F_{com},
\end{equation}where $\odot$ denotes the element-wise multiplication. Then, We pass the relation feature map into a GAP layer and two fully connected layers followed by a softmax layer. In this way, we get the eventual output of the network. 

\textbf{The Training Procedure.}
The cross-entropy loss is used to optimize the network and the training procedure is as follows:

$\bullet$ For the overall architecture in Fig.~\ref{overall_architecture}, the multi-modality fusion module is removed, the two stream sub-networks are made into independent networks by adding a fully connected (FC) layer and a softmax layer on the top.

$\bullet$ These two sub-networks are trained independently and their weights are saved except for the FC and softmax layers.

$\bullet$ Based on the obtained weights, the weights of the two stream sub-networks are fixed in this way and the multi-modality fusion module is trained. Then the whole network is fine-tuned together.

\section{Experiments}
The training datasets and test datasets are introduced in the beginning. Then we list the implementation details in our experiments. Furthermore, both experiments compared with state-of-the-art methods and the ablation study is conducted to test the effectiveness of our proposed modules.

\subsection {Dataset}
We conduct training and testing procedure on two popular video action recognition datasets, NTU RGB+D dataset and SYSU dataset, which are described in detail as follows:

\textbf{NTU RGB+D dataset \cite{ntu_rgbd}:} This dataset contains 60 different action classes in 56880 video samples covering skeleton, depth, IR, and RGB video modalities. In our implementation, we only pick out the skeleton temporal modality and RGB video spatial modality data source. Additionally, 50 classes are performed by a single subject and the rest 10 classes are mutual action performed by two subjects, where both single and double subject actions consist of 25 joints. There are two standard evaluation protocols, Cross-Subject evaluation splitting the 40 subjects into training and testing groups and Cross-View evaluation which utilizes the samples of cameras 2 and 3 for training while samples of camera 1 for testing.
 
\textbf{SYSU dataset \cite{sysu}:} This dataset contains 12 actions performed by 40 subjects with 20 joints. SYSU contains 480 sequences of which all are about human-object interacted actions. There are 6 objects, namely as cellphone, chair, backpack, wallet, cup, broom, mop. There are also two standard evaluation protocol settings: For Setting-1, half of the action sequences are used for training and the rest are used for testing. For Setting-2, half of the subjects are used for training and the rest are used for testing. For each setting mode, 30-fold cross validations are implemented. 

\subsection{Implementation Details}

The model is implemented by PyTorch as the backend. It is trained on four Nvidia GTX 1080Ti GPUs. we chose Adam as the optimizer. The learning rate is set to $10^{-4}$ and reduced by multiplying it by $0.1$ every $10$ epochs. We introduce coordinates transformation to calibrate the dataset. In our implementation, in order to eliminate the influence of the camera sensor position and action position, the pre-processing method VA-pre~\cite{3d_action_joints} is employed. Through this pre-processing method, the original point of the coordinate system is transformed to the body center of the first frame. In the following frames, the coordinate of the skeleton is settled according to the relative location of its center towards the original point.

In the RGB stream, the input image is resized to $299 \times 299$. The skeleton sequences are downsampled to the minimum length of the dataset: 32 frames from NTU RGB+D dataset and 58 frames from SYSU dataset are picked out as the skeleton sequence. For the Bi-LSTM backbone, the dimension of the skeleton feature is set to 150 in NTU RGB+D dataset, thus the dimension of the temporal feature is 600 for bidirectional LSTM. For the ST-GCN backbone, the temporal dimension is set to 300 for both datasets by padding the clips. Thus the output dimension is $256\times 75 \times 25$ for NTU RGB+D dataset and $256\times 75 \times 20$ for SYSU. The dimension of the RGB feature with LSTM based method is $256 \times 3 = 768$ as there are three parts of the same dimensions: two branches for the self-attention feature and one for the skeleton-attention feature. For the RGB frame, we choose the middlemost RGB frame as the representation of the RGB modality.

\subsection{Comparison with State-of-the-Art Methods}
\subsubsection{Comparison of Accuracy} The precision of our model is shown in Table~\ref{tab_sota_ntu} and Table~\ref{tab_sota_sysu} in comparison with most of the recent state-of-the-art methods. 

\textbf{Results on NTU RGB+D.} We compare the Cross-Subject and Cross-View accuracy on NTU RGB+D dataset with SOTA methods including skeleton based methods such as ST-LSTM+Trust Gate~\cite{ntu_rgbd}, VA-LSTM~\cite{adaptive_rnn}, D-Pose Traversal Conv~\cite{dpose}, ST-GCN~\cite{ST-GCN}, DGNN~\cite{DGNN}, MS-G3D Net~\cite{MS-G3D}, RGB based methods such as Glimpse Clouds~\cite{Glimpse_Clouds}, multi-modality methods such as 2 stream RNN/CNN \cite{2streamfusion}, Deep-Bilinear~\cite{bilinear}, Posemap~\cite{posemap},MFAS~\cite{mfas}, SGM-Net~\cite{SGMNet}, MMTM~\cite{MMTM}, JOLO-GCN~\cite{JOLO-GCN}.

\begin{table}[t]
\centering
\caption{Comparison with the state-of-the-art methods on NTU RGB+D. $S$ denotes the skeleton modality, $R$ means RGB modality, and $S+R$ represents utilizing the information from both the modalities.}
\label{tab_sota_ntu}
\begin{tabular}{ccccccc}
\toprule
\multirow{2}{*}{Methods} & \multirow{2}{*}{Type} & \multicolumn{2}{c}{Accuracy} & \multirow{2}{*}{Year} & \multirow{2}{*}{Parameters} & \multirow{2}{*}{FLOPs}\\
 & &Cross-Subject & Cross-View &  &  \\ \hline
VA-LSTM~\cite{adaptive_rnn} & S & 79.4\% & 87.6\% & 2017 & - & - \\
D-Pose Traversal Conv~\cite{dpose} & S& 76.8\% & 84.9\% & 2018 & - & -\\
DPRL+GCNN~\cite{gcnn} & S& 83.5\% & 89.8\% & 2018 & - & -\\
ST-GCN~\cite{ST-GCN} & S& 81.5\% & 88.3\% & 2018 & 3.1M & 16.2G \\
EleAtt-GRU~\cite{EleAtt-GRU} & S & 80.7\% & 88.4\% & 2019 & 0.3M & 6.7G \\
DGNN~\cite{DGNN} & S & 89.9\% & 96.1\% & 2019 & - & - \\
MS-G3D Net~\cite{MS-G3D} & S & 91.5\% & 96.2\% & 2020 & 3.2M & 48.8G \\
SGN~\cite{SGN} & S & 89.0\% & 94.5\% & 2020 & 1.2M & 0.80G \\
Inflated Resnet50~\cite{Glimpse_Clouds}& R & 86.6\% & 93.2\% & 2018 & 46.8M & 168G \\
2D/3D Multitask~\cite{2d3d_pose}& S+R & 85.5\% & -- & 2018 & 12.1M & 107.9G \\
2 Stream RNN/CNN~\cite{2streamfusion} & S+R& 83.7\% & 93.7\% & 2018 & 80.2M &38.6G \\
Deep-Bilinear~\cite{bilinear} & S+R & 83.0\% & 87.1\% & 2018 & - & - \\
Posemap~\cite{posemap} & S+R & 91.7\% & 95.2\% & 2018 & - & -\\
MFAS~\cite{mfas} & S+R& 90.0\% & -- & 2019 & 51.2M &  219.8G \\
SGM-Net~\cite{SGMNet} & S+R& 88.9\% & 95.7\% & 2020 & 71.6M &169.2G \\
MMTM~\cite{MMTM} & S+R& \textbf{91.9\%} & -- & 2020 & 49.4M & 215.5G\\
JOLO-GCN~\cite{JOLO-GCN} & S+R & 90.4\% & 95.8 & 2021 & 6.2M & 20.2G\\
\hline
\textbf{MMFF w/ Bi-LSTM (Ours)} & S+R& 85.4\% & 91.6\% & - & 27.4M & 15.3G\\
\textbf{MMFF w/ ST-GCN (Ours)}& S+R & 89.6\% & \textbf{96.3\%} & - & 29.1M & 24.5G \\ \bottomrule
\end{tabular}
\end{table}

According to the experiment results, GCN based backbones are naturally more powerful than RNN based backbones due to their higher ability in reasoning the regional relation among the movement of human limbs. It is shown that MMFM outperforms the single modality methods significantly with minimal increase of computational workload. For example, MMFF achieves $4.7\%$ better on Cross-Subject evaluations than EleAtt-GRU~\cite{EleAtt-GRU} with only 8.6G more FLOPs. 

To be fair, we focus to compare the proposed MMFF with the methods which use the same backbone for skeleton stream. Compared with 2 Stream RNN/CNN~\cite{2streamfusion} with Bi-LSTM as the backbone, MMFF with Bi-LSTM achieves competitive results. As to SGM-Net~\cite{SGMNet} with ST-GCN as the backbone, MMFF with ST-GCN has superior performance on both evaluations with far fewer parameter and FLOPs. Such performance reveals that though SGM-Net utilizes both skeleton sequence and whole RGB video modalities, our MMFF model using only a single frame with two stage feature fusion still exceeds it in both performance and computational consumption. Also, in comparison with the recent fusion methods MMTM~\cite{MMTM} and MFAS~\cite{mfas} which use Inflated ResNet50~\cite{Glimpse_Clouds} for video processing and HCN~\cite{HCN} for skeleton processing, MMFF achieves competitive results with only $11.4\%$ of the FLOPs of MMTM and $11.1\%$ of the FLOPs of MFAS. As to Posemap~\cite{posemap} and JOLO-GCN~\cite{JOLO-GCN} whose result with the same backbone ST-GCN is reported, MMTM achieved competitive results. For Posemap, the parameter and FLOPs are not reported in the paper. Since Posemap uses CNN to extract the features of RGB video, the computational workload can be relatively large. JOLO-GCN proposes constructing the optical flow motions around each joint to form a graph and using the lightweight GCN to extract the features. Even though the parameter and FLOPs of JOLO-GCN are fewer than the proposed MMFF, the consumption of processing the image to optical flow and the joint guided optical flow graphs is not considered.

\textbf{Results on SYSU.} As to SYSU dataset, We compare both Setting-1 and Setting-2 on SYSU dataset with methods including the skeleton-based methods VA-LSTM~\cite{adaptive_rnn}, DPRL+GCNN~\cite{gcnn}, Local+LGN~\cite{Local_LGN}, EleAtt-GRU~\cite{EleAtt-GRU}, LSGM+GTSC~\cite{LSGM+GTSC}, SGN~\cite{SGN} and the fusion methods MTDA~\cite{mtda}, JOULE~\cite{joule}, Deep-Bilinear~\cite{bilinear}, PI3D~\cite{PI3D}. It can be observed that the proposed methods with both Bi-LSTM and ST-GCN achieve competitive performance. Specifically, MMFF with ST-GCN outperforms the multi-modality fusion method JOULE~\cite{joule} by $2.2\%$ in Setting-2 on SYSU with only $12.0\%$ of the FLOPs of JOULE. When compared with PI3D~\cite{PI3D}, which uses I3D~\cite{I3D} for video processing, MMFF also achieves superior performance on setting-2 evaluation. Though EleAtt-GRU~\cite{EleAtt-GRU} achieves SOTA performance on Setting-1 evaluation, it is pre-trained on NTU RGB+D.

\begin{table}[t]
\centering
\caption{Comparison with state-of-the-art methods on SYSU. $S$ denotes the skeleton modality, $R$ means RGB modality, and $S+R$ represents utilizing the information from both the modalities.}
\label{tab_sota_sysu}
\begin{tabular}{ccccccc}
\toprule
\multirow{2}{*}{Methods} & \multirow{2}{*}{Type} & \multicolumn{2}{c}{Accuracy} & \multirow{2}{*}{Tear} & \multirow{2}{*}{Parameters} & \multirow{2}{*}{FLOPs} \\
 & & Setting-1 & Setting-2 &  &  \\ \hline
VA-LSTM~\cite{adaptive_rnn} & S & 76.9\% & 77.5\% & 2017 & - & - \\
DPRL+GCNN~\cite{gcnn} & S & 76.9\% & - & 2018 & - & - \\
Local+LGN~\cite{Local_LGN} & S & 83.1\% & - & 2019 & - & - \\
EleAtt-GRU~\cite{EleAtt-GRU} & S & \textbf{85.7}\% & 85.7\% & 2019 & 0.3M & 6.7G \\
LSGM+GTSC~\cite{LSGM+GTSC} & S & - & 85.8\% & 2020 & - & -\\
SGN~\cite{SGN} & S & 81.6\% & 83.0\% & 2020 & 1.2M & 0.8G\\
MTDA~\cite{mtda}& S+R & 79.2 \% & 84.5\% & 2011 & 46.6M & 202.9G \\
JOULE~\cite{joule}& S+R & 79.6\% & 84.9\% & 2017 & 47.4M & 203.8G\\
Deep-Bilinear~\cite{bilinear}& S+R & 81.5\% & 86.2\% & 2018 & - & - \\
PI3D~\cite{PI3D} & S+R & - & 85.8\% & 2021 & - & -\\
\hline
\textbf{MMFF w/ Bi-LSTM (Ours)} & S+R & 80.9\% & 82.6\% & - & 27.4M & 15.3G \\
\textbf{MMFF w/ ST-GCN (Ours)} & S+R & 85.3\% &  \textbf{87.1\%} & - & 29.1M & 24.5G \\ \bottomrule
\end{tabular}
\end{table}


\subsubsection{Comparison of Complexity.} Current multi-modality methods either apply LSTM based or GCN based networks as the backbone to process the skeleton modality. Typically, multiple Bi-LSTMs~\cite{BiLSTM} (2.0M parameters and 6.3G FLOPs) achieves the best performance in LSTM based methods and ST-GCN~\cite{ST-GCN} (3.1M parameters and 15.6G FLOPs) is of the most popularity in GCN based networks. Our MMFF method also utilizes the two backbones mentioned above to extract the temporal information from skeleton sequences. Thus the complexity in the skeleton stream is similar among the current methods.
As to the RGB stream, the state-of-the-art fusion methods perform either Inflated ResNet50~\cite{Glimpse_Clouds} with 46.8M parameters and 168G FLOPs or C3D~\cite{C3D} with 12.1M parameters and 107.9G FLOPs to obtain the spatial information. However, our MMFF extracts the spatial representation from a single RGB frame by Xception network with 22.86M parameters and 8.42G FLOPs, which significantly saves the computational consumption. For the fusion stage, most of the fusion methods only require 2-3 M parameters, the difference of which can be ignored. In conclusion, our method exceeds other state-of-the-methods in efficiency mostly thanks to the choice of the single RGB frame.

For example, when compared with 2 stream RNN/CNN~\cite{2streamfusion}, our method performs much better and uses only $32.6\%$ and $63.5\%$ of the parameters and FLOPs of 2 stream RNN/CNN. Also, Our MMFF method has a competitive performance with MMTM~\cite{MMTM} which uses ST-GCN and I-ResNet50 as the backbone with 49.4M parameters and 215.5G FLOPs, while our ST-GCN based model only requires 29.1M parameters ($49.6\%$ of MMTM) and 24.5G FLOPs ($11.4\%$ of MMTM). As to the SOTA method PI3D~\cite{PI3D}, MMFF has the advantage of low complexity, as PI3D uses I3D~\cite{I3D} for RGB video processing, where the parameter and FLOPs of I3D are 12.2M and 55.9G. Therefore, our method achieves competitive performance with the state-of-the-art methods but saves the computational consumption significantly.

\subsection{Ablation Study}
In this section, we introduce the ablation studies to test the methods effectiveness on both NTU RGB+D and SYSU datasets. The ablation studies include the choice of RGB frame selection period, the choice of the number of frames, the effectiveness of data enhancement, the effectiveness of self-attention and the effectiveness of two-stage fusion methods and the backbone in processing skeleton sequence.
\subsubsection{The Choice of RGB Frame Time Period.} 
\begin{figure}[t]
    \centering
    \includegraphics[width =4.0in]{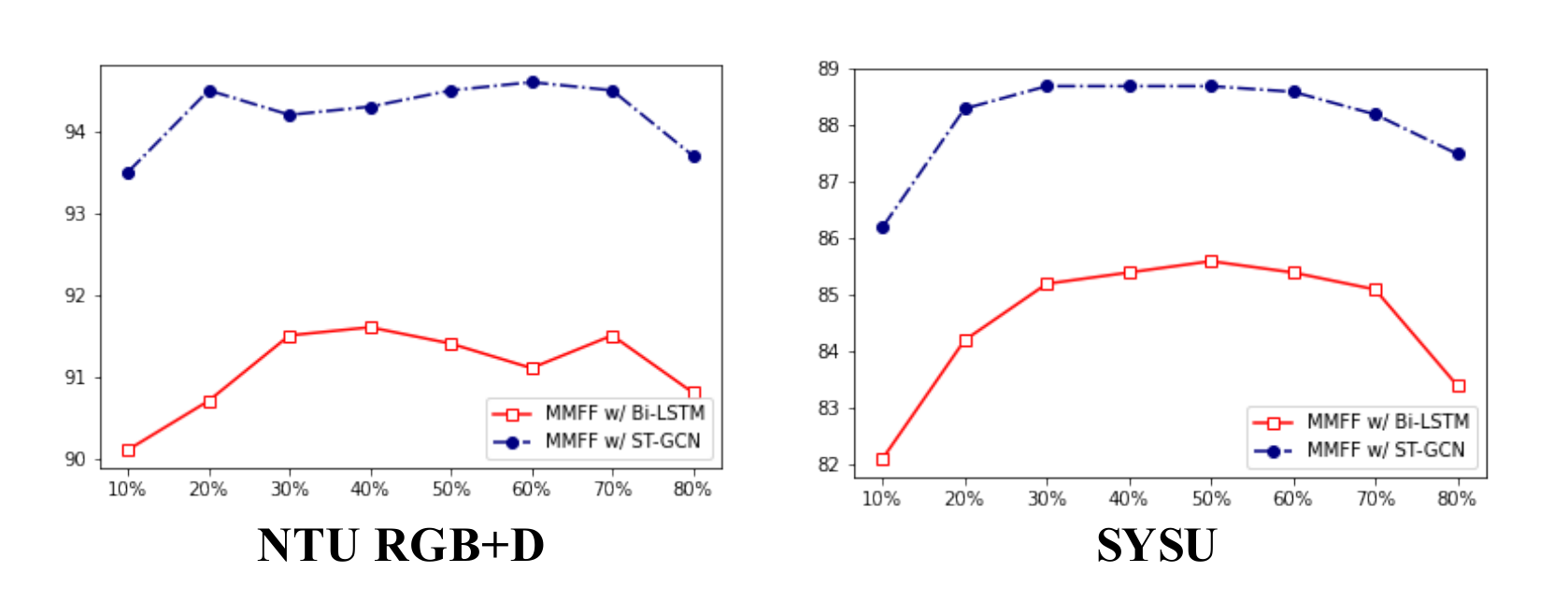}
    \caption{The ablation study on SYSU and NTU RGB+D between the choice of LSTM and ST-GCN in the implementation of skeleton temporal sequence. Horizontal axis denotes the time period percentage in which the single RGB frame is extracted and vertical axis denotes the performance.}
    \label{fig_frame_gcn}
\end{figure}

In the RGB stream, we pick out a single RGB frame instead of the whole video clip for efficiency. We conduct the ablation experiment to test the frame of which time period is the most suitable to choose with both LSTM and ST-GCN backbones on NTU RGB+D and SYSU. The percentage of the time period ranges from 10\%, 20\%, 30\%,... to 90\%. The result of the precision of the feature fusion module with each different RGB frame is listed in Fig.~\ref{fig_frame_gcn}. The result shows that the frames from the middle part of the RGB video achieve similar performance. Because the difference in the choice of frames between $30\%$ to $70\%$ is minimal according to the experiments, thus the choice of the single RGB frame is very flexible, which proves our assumption that a single RGB frame with the human-object interaction covers enough spatial information. To unify the standard for convenience, we extract the $50\%$ middle RGB frame as the RGB feature representation.
\subsubsection{The Choice of the Number of Frames}
We conduct experiments on NTU RGB+D to figure out the relation among the number of frames, performance and parameters. The reason why we choose the frames located at $30\%, 40\%, 50\%, 60\%, 70\%$ of the video is that Figure~\ref{fig_frame_gcn} reveals that the model achieves the best performance with the frames picked out in this region. It is shown that the performance increases slightly with the addition of frames, while the parameters increase dramatically with the multiple uses of the Xception~\cite{Xception} network. Thus we decide to only use a single frame to balance the efficiency. Unlike~\cite{how_many} which only learns temporal information from RGB modality, we can learn temporal information from the skeleton stream, thus it is better for us to pick no more than one frames, considering the computational consumption.

\begin{table}[t]
\caption{Different number of frames are picked out from the video and combined with MMFF. }
\begin{tabular}{@{}ccccc@{}}
\toprule
\multirow{2}{*}{Frames} & \multirow{2}{*}{Position} & \multicolumn{2}{c}{NTU RGB+D} & \multirow{2}{*}{Parameters} \\
 &  & Cross-Subject & Cross-View &  \\ \midrule
1 & 50\% & 89.6\% & 96.3\% & 29.1M \\
3 & 30\%, 50\%, 70\% & 90.1\% & 96.6\% & 90.1M \\
5 & 30\%, 40\%, 50\%, 60\%, 70\% & 90.3\% & 96.3\% & 134.9M \\ \bottomrule
\end{tabular}

\label{choice_number}
\end{table}

\subsubsection{The Effectiveness of Data Enhancement.} Data enhancement consists of the data augmentation and projection crop, which are used for skeleton and RGB, respectively. We conduct experiments with ST-GCN \cite{ST-GCN} as the backbone on both NTU RGB+D and SYSU datasets to test the effectiveness of our proposed data enhancement technique. For the methods Xception \cite{Xception}, MFAS \cite{mfas} and MMFF without Data Enhancement, the same commonly used crop method is applied to them, which randomly crops the video to a size of $224\times 224$ patch.

\begin{table}[t]\footnotesize
\centering
\caption{Ablation experiments on the implementation of data enhancement, in which data augmentation is applied on skeleton sequence modality and data augmentation is performed on RGB image modality. Xception denotes the baseline network which processes the single RGB frame. In the last row, the data enhancement are added to the whole network. We use the code of MFAS for training and testing the Cross-View evaluation.}
\label{tab_data_enhancement}
\begin{tabular}{ccccc}
\toprule
\multirow{2}{*}{Methods}                                                    & \multicolumn{2}{c}{NTU RGB+D}      & \multicolumn{2}{c}{SYSU}          \\
& Cross-Subject   & Cross-View      & Setting-1       & Setting-2       \\ \midrule
ST-GCN & 81.5\% & 88.3\%   & 78.6\% &  79.8\%  \\
ST-GCN w/ Data Augmentation & 82.3\% & 89.6\%  & 80.4\% &  81.2\% \\
\hline
Xception & 49.9\% & 50.2\% & 52.2\% & 55.4\% \\
Xception w/ Projection Crop & 65.4\% & 70.8\% & 62.4\% & 65.7\% \\
\hline
MFAS~\cite{mfas} & 90.0\% & 93.2\% & - & - \\
MFAS w/ Data Augmentation & 90.8\% & 94.7\% & - & - \\ \hline
MMFF w/o Data Enhancement& 83.0\% & 90.8\% & 80.6\% & 81.4\% \\
MMFF w/o Data Augmentation & 88.9\% & 95.2\% & 83.7\% & 85.1\% \\
MMFF w/o Projection Crop & 85.3\% & 93.4\% & 84.3\% & 85.6\% \\
MMFF & 89.6\% & 96.3\% & 85.3\% & 87.1\% \\
\bottomrule
\end{tabular}
\end{table}

As illustrated in Table~\ref{tab_data_enhancement}, the data enhancement increases the accuracy of our network reasonably. In detail, for the skeleton modality, the data augmentation helps the baseline ST-GCN \cite{ST-GCN} improve $0.8\%$ in Cross-Subject on NTU RGB+D, showing that providing different viewpoint augmentation does help the model achieve more robust and satisfying results. Also, in the RGB stream only, we test the effectiveness of ProjCrop processed by the Xception network. It is shown that ProjCrop boosts the performance of RGB image by $10.2\%$ in Setting-1 on SYSU, which further proves our assumption that too much scattered background may interfere the performance of the RGB frame model and the crop on the frames can help the model focus more on human bodies to some extent. When added to the whole network MMFF, the data enhancement consistently helps improve the performance. 

To further evaluate the effects of data enhancement, we conduct experiments on MFAS \cite{mfas} with data augmentation for skeleton sequence modality. It is indicated that data augmentation helps improve the performance over the original MFAS on NTU RGB+D dataset. Along with the performance comparison on ST-GCN, data augmentation proves to be effective for current skeleton-based methods. Moreover, we test MMFF without data augmentation and MMFF without projection crop for a fair comparison with MFAS. MMFF without data augmentation still has competitive performance with MFAS. It can also be concluded that MMFF benefits more from projection crop on NTU RGB+D dataset and more from data augmentation on SYSU dataset. The reason is that projection crop can help reduce the background area of images in NTU RGB+D dataset and data augmentation can increase the training data for the relatively small dataset to avoid overfitting.

\subsubsection{Effectiveness of Self-Attention Mechanism.} We conduct experiments with Bi-LSTM and ST-GCN based backbones on both NTU RGB+D and SYSU datasets to test the effectiveness of the self-attention mechanism.

\begin{table}[t]\footnotesize
\centering
\caption{Performance of the combination of the modules on Bi-LSTM and ST-GCN based backbones. Self-Attention denotes the self-attention module and Skeleton-Attention stands for the skeleton attention module. MMFF means the utilization of both self and skeleton attention modules. Decision Fusion is the naive weighted sum decision fusion method and Sum Fusion is to sum the outputs of Bi-LSTM-based method following with fully-connected layers and softmax for classification.}
\label{tab_ablation}
\renewcommand\tabcolsep{2pt}
\begin{tabular}{c|ccccc}
\toprule
\multirow{2}{*}{Backbone} & \multirow{2}{*}{Methods} & \multicolumn{2}{c}{NTU RGB+D} & \multicolumn{2}{c}{SYSU} \\
& & Cross-Subject & Cross-View & Setting-1 & Setting-2 \\ \midrule
\multirow{6}{*}{Bi-LSTM} & Baseline & 72.3\% & 79.4\% & 72.3\% & 72.8\%  \\
& MMFF w/o Skeleton-Attention & 80.3\% & 85.5\% & 75.7\% & 77.5\% \\
& MMFF w/o Self-Attention & 81.2\% & 87.4\% & 77.3\% & 78.4\% \\
& MMFF-Decision Fusion & 81.0\% & 86.9\% & \textbf{90.3\%} & 81.2\% \\
& MMFF-Sum Fusion & 83.0\% & 89.9\% & 80.3\% & 80.9\% \\
& MMFF & \textbf{85.4\%} & \textbf{91.6\%} & 80.9\% & \textbf{82.6\%} \\  \midrule
\multirow{5}{*}{ST-GCN} & Baseline & 81.5\% & 88.3\% & 78.6\% & 79.8\%\\
& MMFF w/o Skeleton-Attention & 84.2\% & 92.5\% & 82.0\% & 83.3\% \\
& MMFF w/o Self-Attention & 86.4\% & 94.1\% & 82.2\% & 84.6\%\\
& MMFF-Decision Fusion & 87.2\% & 94.4\% & 84.5\% & 85.1\%\\
& MMFF & \textbf{89.6\%} & \textbf{96.3\%} & \textbf{85.3}\% & \textbf{87.1}\%\\ \bottomrule
\end{tabular}
\end{table}

As shown in Table~\ref{tab_ablation}, the self-attention mechanism brings prominent improvement to the backbone network. For example, the self-attention module improves the performance of MMFF with Bi-LSTM w/o self-attention by $1.8\%$ in Cross-Subject on SYSU,  which proves that our proposed self-attention module helps the RGB stream concentrate more on the human-object interaction regions to extract the more representative information. Also, skeleton attention improves the MMFF with ST-GCN w/o skeleton attention by $3.0\%$ in Cross-Subject on NTU RGB+D, revealing that the skeleton attention can effectively help better fuse the two modalities as well as help the model concentrate more on telling the foreground from the background. 

\subsubsection{Effectiveness of the Early Fusion Module.} We carry out experiments with Bi-LSTM and ST-GCN based network to prove the performance of our proposed early fusion method. The skeleton attention mechanism is used as the early fusion stage to transfer the temporal knowledge to the RGB stream. As described in
Table~\ref{tab_ablation}, skeleton attention improves Bi-LSTM by $13.5\%$ in Setting-1 on SYSU and brings $8.0\%$ increase in Cross-View on NTU RGB+D. The data underpins our proposal that the projection attention mechanism of the skeleton sequence not only guides the RGB frame in focusing on the human-object interaction region but also helps further improve the feature fusion by serving as the early stage of fusion. Moreover, we can also observe that the skeleton-attention plays a more important role than the self-attention.

\subsubsection{Effectiveness of the Late Fusion Module.} In our implementation, we test the late fusion methods based on LSTM and ST-GCN backbones on NTU RGB+D and SYSU datasets. We compare the performance of the proposed methods with the decision fusion and sum fusion methods. The results are reported in Table~\ref{tab_ablation}.

\begin{figure}[t]
    \centering
    \includegraphics[width=\textwidth]{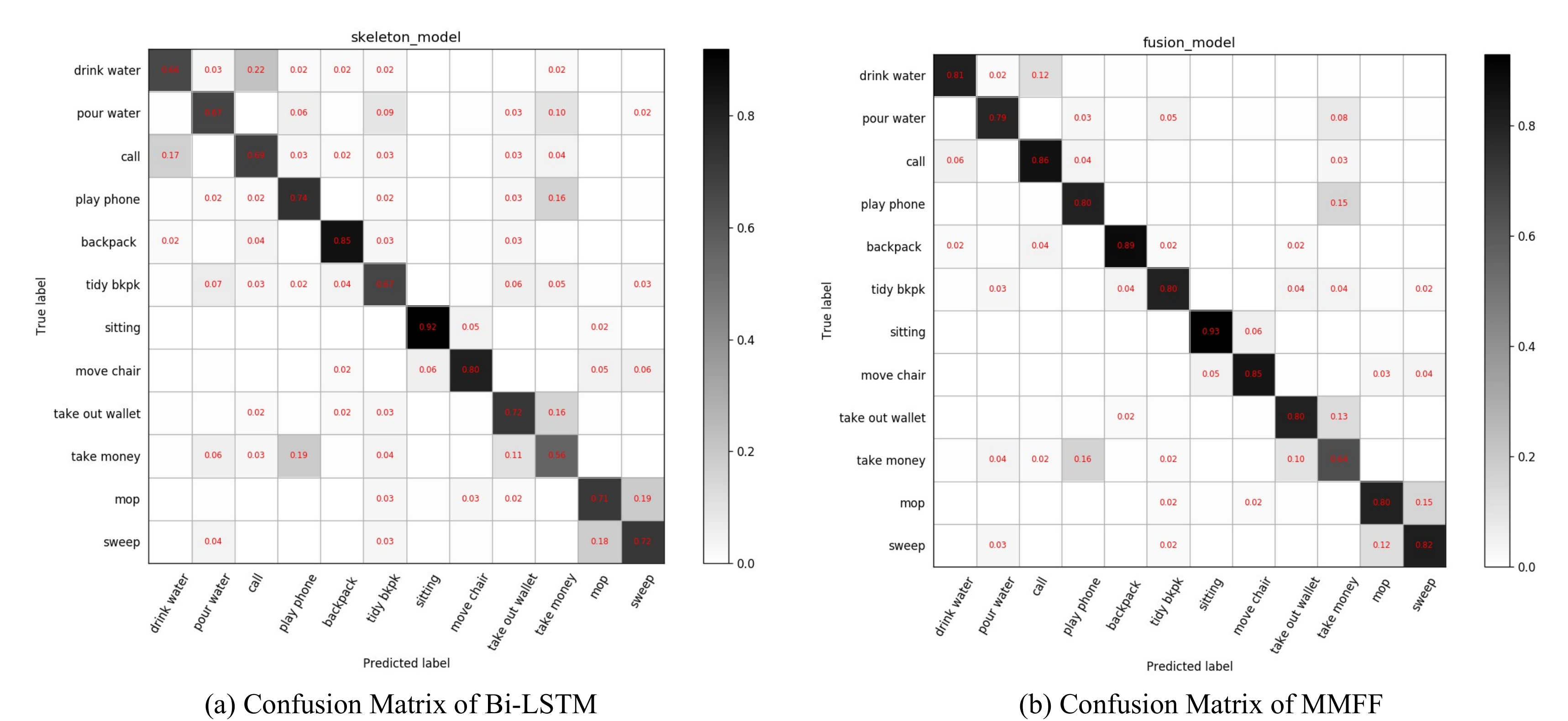}
    \caption{The confusion matrices of SYSU dataset: (a) The result of Bi-LSTM model (b) The result of MMFF model. The results are the sum of Setting-1 and Setting-2 for all 30-fold cross validation.}
    \label{fig_confusion_matrix}
    
\end{figure}

It is shown that the precision of MMFF outperforms the other methods on both NTU RGB+D and SYSU datasets. For both the Bi-LSTM and ST-GCN based method, the proposed late fusion module achieves consistent improvement over most results of the decision fusion method. For example, our proposed late fusion module on Bi-LSTM achieves higher performance compared with the decision fusion method by $5.0\%$ in Cross-View on NTU RGB+D and $3.8\%$ in Setting-1 on SYSU. However, the best performance on SYSU Setting-1 evaluation is achieved with decision fusion. Considering the fact that decision fusion does not generalize well on larger NTU RGB+D dataset, the result can be regarded as abnormal. Moreover, the MMFF with the proposed late fusion module also exceeds the sum fusion method by $2.4\%$ in Cross-Subject on NTU RGB+D and $2.0\%$ in Setting-1 on SYSU.

\subsubsection{Accuracy Analysis on Categories.}
Fig.~\ref{fig_confusion_matrix} describes the effectiveness of Bi-LSTM based MMFF with confusion matrices on SYSU dataset. The element in the matrix denotes the probability of classifying one kind of action to others and the elements on the diagonal denote the probability of correct classification. According to the fusion matrices, we observe that the diagonal elements of MMFF improve consistently over those of the backbone Bi-LSTM, which indicates that MMFF helps boost the performance of different classes. As the actions in SYSU mainly contain human-object interaction, the proposed method benefits to fuse the spatial information for action recognition.

\begin{figure}[]
    \centering
    \includegraphics[width=0.9\textwidth]{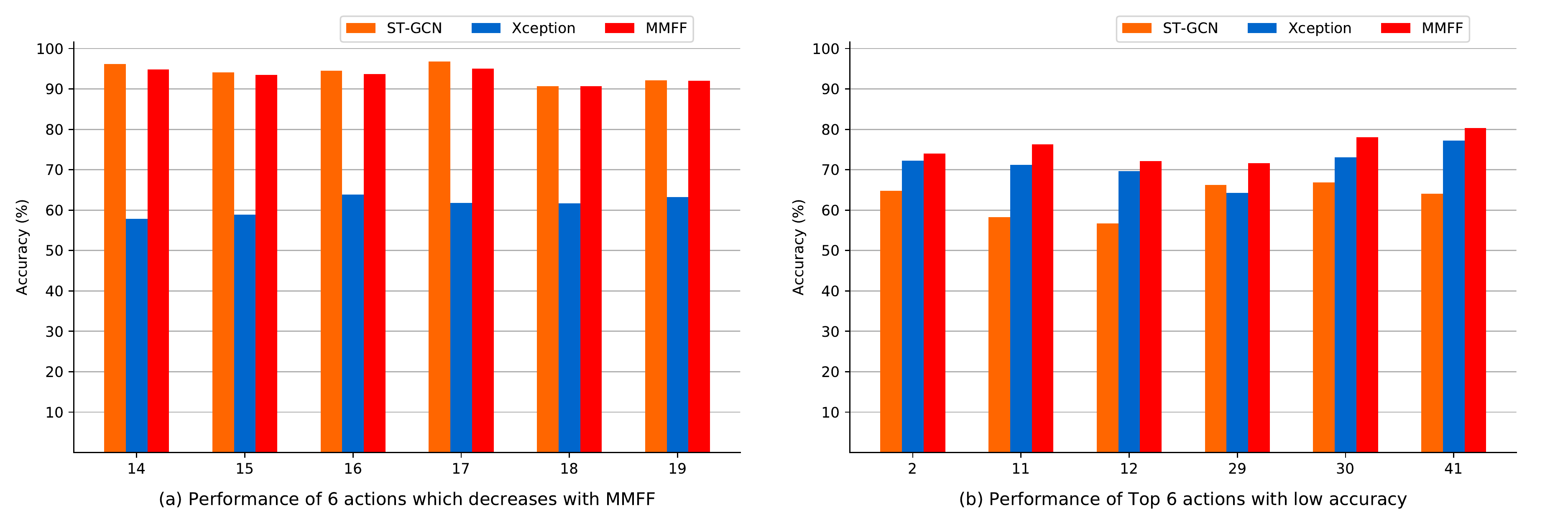}
    \caption{Performance of typical actions on NTU RGB+D Cross-Subject. The numbers on the horizontal axis denote the action labels. The actions are "wear jacket" (14), "take off jacket" (15), "wear a shoe" (16), "take off a shoe" (17), "wear on glasses" (18), "take off glasses" (19), "eat meal/snack" (2), "reading" (11), "writing" (12), "playing with phone/tablet" (29), "typing on a keyboard" (30), "sneeze/cough" (41).}
    \label{NTU_case}
\end{figure}

As to NTU RGB+D dataset, we focus on analyzing the categories with performance drop and low accuracy, while MMFF achieves consistent improvement on the other actions. As shown in Fig. \ref{NTU_case}(a), we observe that the performance of six actions slightly decreases, for Xception has difficulty classifying the actions with the same objects and moving regions (e.g., wear jacket and take off jacket). As to the challenging actions for ST-GCN, MMFF can improve the performance with the help of the RGB image modality.

\subsubsection{The Choice of the Backbone in Processing Skeleton Sequence.} We conduct experiments on NTU RGB+D and SYSU datasets to determine the proper backbone for LSTM based network. As shown in Table~\ref{tab_layers_lstm}, triple Bi-LSTM achieves the highest performance in LSTM based methods. However, as GCN based methods have achieved much better performance than LSTM based methods, to be fair, both the LSTM based and GCN based backbones are used in comparison with the state-of-the-art methods.

\begin{table}[t]
\centering
\caption{Performance of Proposed Method with LSTM-based backbones on NTU RGB+D and SYSU datasets.}
\label{tab_layers_lstm}
\begin{tabular}{ccccc}
\toprule
     \multirow{2}{*}{Methods}                   & \multicolumn{2}{c}{NTU RGB+D}      & \multicolumn{2}{c}{SYSU}          \\
                        & Cross-Subject   & Cross-View      & Setting-1       & Setting-2       \\ \midrule

single Bi-GRU           & 66.9\%          & 73.8\%          & 70.7\%          & 71.1\%          \\
double Bi-GRU           & 69.8\%          & 77.6\%          & 71.6\%          & 72.0\%          \\
triple Bi-GRU           & 72.1\%          & 78.9\%          & 72.1\%          & 72.6\%          \\ \hline
single LSTM              & 64.6\%          & 70.2\%          & 62.1\%          & 63.2\%          \\
double LSTM             & 66.8\%          & 72.1\%          & 64.9\%          & 65.8\%          \\
triple LSTM             & 67.0\%          & 72.1\%          & 64.9\%          & 65.8\%          \\ \hline
single Bi-LSTM          & 67.7\%          & 73.7\%          & 69.3\%            & 70.9\%          \\
double Bi-LSTM          & 70.6\%          & 76.5\%          & 71.9\%          & 72.4\%          \\
\textbf{triple Bi-LSTM} & \textbf{72.3\%} & \textbf{79.4\%} & \textbf{72.3\%} & \textbf{72.8\%} \\ \bottomrule
\end{tabular}
\end{table}

\begin{figure}[t]
    \centering
 
    \includegraphics[width =3.5in]{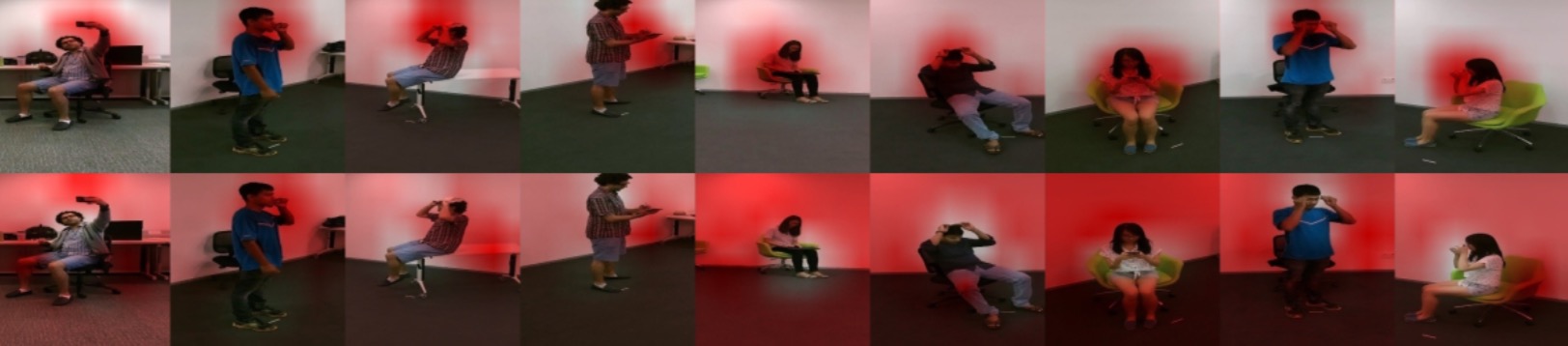}
    \caption{The heatmap of the self-attention. The first column tends to focus on body part while the second column focuses on the background.}
    \label{fig_self_attention}
    
\end{figure}

\begin{figure}[t]
    \centering
 
    \includegraphics[width =3.5in]{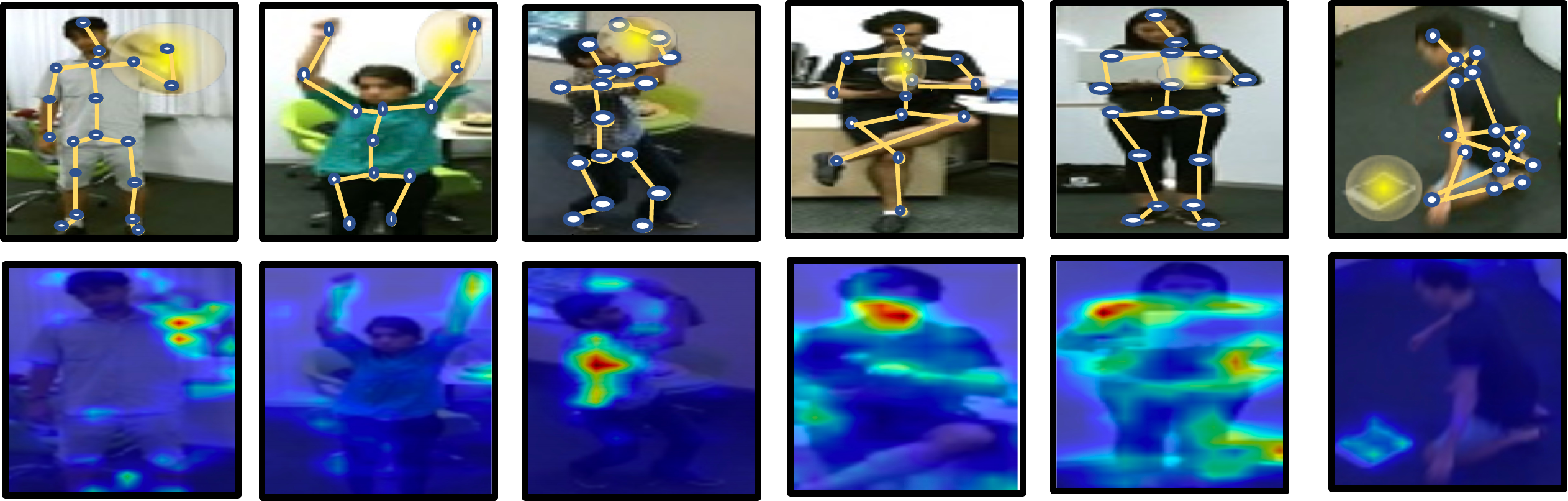}
    \caption{The illustration of the skeleton attention mechanism. It is shown that the implemented skeleton attention mechanism help the network focus more on the body movement part. }
    \label{skeleton_attention}
    
\end{figure}

\subsection{Visualization}
The figure of self-attention module and skeleton-attention module are described in Fig.~\ref{fig_self_attention} and Fig.~\ref{skeleton_attention}, respectively. All of the pictures are cropped as described in the implementation details. It is shown that the self-attention helps the model focus more on the human-object interaction region, while the skeleton-attention helps the model focus more on the movement of the human limbs.

\section{Discussion}
\subsection{Adaptation to Different Datasets} The proposed MMFF achieves consistent performance on both SYSU and NTU RGB+D datasets with lower complexity, yet the method shows different dataset properties. As SYSU dataset is relatively small, the network benefits more from data augmentation. As to NTU RGB+D datasets with more challenging actions, introducing the RGB frame modality with MMFF helps boost the performance. Even though some single modality methods achieve competitive performance, they do not perform well under other evaluations (e.g., SGN \cite{SGN}), which illustrates the importance of complementary multi-modalities especially for complicated actions recognition.

\subsection{Convenience and Limitation of MMFF} 
The main advantage of MMFF is the convenience of training and inference. Compared with the SOTA multi-modality methods, MMFF achieves better or comparable performance with much lower complexity. Notably, although the complexity of JOLO-GCN is low, it requires 1.5s to pre-process the optical flow for each video \cite{JOLO-GCN}, while the time of MMFF for pre-processing can be negligible. The main limitation is that MMFF does not show obvious superiority over the SOTA skeleton-based methods which develop lightweight GCNs by exploiting motion and temporal correlations. However, we contribute to propose a fusion method by substituting RGB video with RGB frame for simplicity, which is easy to corporate with other skeleton-based methods. The baseline GCN model is used for a fair comparison with the multi-modality methods and the more effective GCN model can be combined with the proposed method for performance improvement.

\section{Conclusion}
This article proposes a two stage multi-modality feature fusion model for action recognition. Our main contribution is to extract one RGB frame of the middle part of the whole video to maintain most of the key information from the RGB stream as well as increase the efficiency. To better learn the correspondence of the two modalities, we apply the skeleton-attention module as the early fusion stage, which transfer the temporal knowledge from the skeleton modality to the RGB modality to help the model focus more on the movement region of human limbs on the RGB frame. As to the late fusion stage, we introduce a fusion network to better fuse the two modalities. Experiments on both LSTM and ST-GCN backbones are performed to test the effectiveness, of which the results demonstrate that the proposed method obtains better or competitive results than state-of-the-art methods and reduces the complexity. Nevertheless, the work may fall in difficulty in some particular applications such as live stream video surveillance, thus we may follow the work and try novel methods to apply the MMFF into these applications.


%






%

\bibliographystyle{ACM-Reference-Format.bst}
\bibliography{main.bib}

\end{document}